\documentclass{article}

\usepackage[final]{aaj2026}
\usepackage[numbers]{natbib}

\usepackage{listings}

\lstdefinestyle{plaincode}{
  basicstyle=\ttfamily\small,
  breaklines=true,
  columns=fullflexible,
  keepspaces=true,
  showstringspaces=false,
  frame=none
}

\usepackage[utf8]{inputenc} 
\usepackage[T1]{fontenc}    
\usepackage{hyperref}       
\usepackage{url}            
\usepackage{booktabs}       
\usepackage{amsfonts}       
\usepackage{nicefrac}       
\usepackage{microtype}      
\usepackage[dvipsnames]{xcolor}
\usepackage[many]{tcolorbox}
\usepackage{multirow}
\usepackage{booktabs}

\title{Evaluation of LLMs in retrieving food and nutritional context for RAG systems}

\author{%
  Maks Požarnik Vavken \\
  Department of Computer Systems\\
  Institute "Jožef Stefan"\\
  Ljubljana, Slovenija \\
  \texttt{mp5628@student.uni-lj.si} \\
\And
 Matevž Ogrinc\\
  Department of Computer Systems\\
  Institute "Jožef Stefan"\\
  Ljubljana, Slovenija \\
  \texttt{matevz.ogrinc@ijs.si} \\
\And
 Tome Eftimov\\
  Department of Computer Systems\\
  Institute "Jožef Stefan"\\
  Ljubljana, Slovenija \\
  \texttt{tome.eftimov@ijs.si} \\  
\And
 Barbara Koroušić Seljak\\
  Department of Computer Systems\\
  Institute "Jožef Stefan"\\
  Ljubljana, Slovenija \\
  \texttt{barbara.korousic@ijs.si} \\  
}

\begin{document}

\maketitle
\begin{abstract}
  In this article, we evaluate four Large Language Models (LLMs) and their effectiveness at retrieving data within a specialized Retrieval-Augmented Generation (RAG) system, using a comprehensive food composition database. Our method is focused on the LLMs' ability to translate natural language queries into structured metadata filters, enabling efficient retrieval via a Chroma vector database. By achieving high accuracy in this critical retrieval step, we demonstrate that LLMs can serve as an accessible, high-performance tool, drastically reducing the manual effort and technical expertise previously required for domain experts, such as food compilers and nutritionists, to leverage complex food and nutrition data. However, despite the high performance on easy and moderately complex queries, our analysis of difficult questions reveals that reliable retrieval remains challenging when queries involve non-expressible constraints. These findings demonstrate that LLM-driven metadata filtering excels when constraints can be explicitly expressed, but struggles when queries exceed the representational scope of the metadata format.
\end{abstract}

\section{Introduction}

The increasing volume and complexity of food and nutrition data challenge current database and knowledge base management systems, which fail to offer domain experts an easy access to integrated, multidimensional information. Many existing systems lack sufficient data granularity, completeness, and interactivity, limiting their practical utility for domain experts without technical skills \cite{zhou2022food}. Similarly, dietitians described current digital tools as outdated, impersonal and poorly adapted to local contexts \cite{manaf2024exploring}. Together, these findings reveal that current systems often fail to supply data based on practitioner needs.

To bridge this gap, Retrieval-Augmented Generation (RAG) systems offer a promising alternative by enabling direct natural language querying of databases \cite{9458778}. In this study, we evaluate how reliably this paradigm operates when applied to food composition data. One critical component of a RAG system is its ability to accurately and reliably retrieve context from a specialized database. Unlike traditional approaches, our system uses a Large Language Model (LLM) to transform a natural language query, such as "Which foods have more than 12g of protein?" into the structured metadata filters required to efficiently query the underlying Chroma \cite{chroma} vector database which we use to store and retrieve semantic embeddings, created from Slovenian food composition data and its corresponding metadata. To ensure accessibility for non-technical users, this retrieval mechanism is exposed through a simple user interface that allows food compilers and nutritionists to interact with the system using natural language.

This paper focuses on evaluating the critical step of context retrieval accuracy of three proprietary LLMs and one commonly used open-access LLM (Gemini, GPT, Claude, Mistral) in generating the necessary structured queries for data extraction in a RAG system.
\section{Related Work}
RAG is a technique that enhances the capabilities of general LLMs by integrating external information with retrieval mechanisms \cite{lewis2020retrieval}. This approach allows LLMs to access and incorporate up-to-date, domain-specific knowledge, improving the accuracy and relevance of generated responses. In healthcare and nutrition, RAG has been instrumental in addressing challenges such as information accuracy and the extraction of complex dietary information.

A comprehensive review by Amugongo et al. (2025) \cite{10.1371/journal.pdig.0000877} assessed RAG-based approaches employed by LLMs in healthcare, focusing on the different steps of retrieval, augmentation, and generation. The study identified various datasets, methodologies, and evaluation frameworks, highlighting the strengths and limitations of existing solutions in this domain.

In parallel to RAG research, a closely related line of work focuses on translating natural language instructions into structured database queries, commonly referred to as Text-to-SQL or NL2SQL. Recent research highlights that LLMs are increasingly capable of generating accurate SQL queries when provided with schema-aware prompts \cite{shi2024survey}. Approaches such as DIN-SQL \cite{pourreza2023din} demonstrate that LLMs can decompose complex conditions into interpretable components, enabling robust structured retrieval over tabular data. Our method (described in detail in Section III) aligns with this paradigm, although our metadata filter format differs from SQL syntax, the underlying task is analogous to NL2SQL, mapping natural language constraints into executable query logic. This connection motivates future exploration of prompt optimization and error correction strategies proposed in the NL2SQL literature to further improve metadata filter generation accuracy.

\subsection{Applications of RAG and LLMs in Food and Nutrition}
The integration of RAG and LLMs has led to significant advancements in dietary information extraction and recommendations. These technologies enable the development of systems that can understand complex dietary queries, provide evidence-based recommendations, and support clinical decision-making.

For instance, the NutriRAG \cite{zhou2025nutrirag} framework utilizes RAG with LLMs to classify food items into predefined food categories, demonstrating the effectiveness of this approach in the food domain.

In~\cite{10.3389/fnut.2025.1635682}, the authors developed an AI-driven dietary recommendation system that generates recipes for patients with obesity and type 2 diabetes. The system prioritizes health outcomes and environmental sustainability, showcasing the potential of RAG based LLM systems in nutrition.

Similarly, Sha et al. (2025) \cite{info:doi/10.2196/75279} integrated a knowledge graph with a RAG system for true/false or multiple-choice question answering related to dietary supplements and drug interactions. Their approach achieved higher accuracy than standalone LLMs, highlighting the benefits of combining RAG with structured knowledge sources.

Beyond recommendation and QA systems, recent work has explored the adaptation of LLMs for domain-specific reasoning and entity grounding. The FoodSEM framework \cite{gjorgjevikj2025foodsem} fine-tunes language models for food-domain named entity linking by aligning food mentions with structured food concepts. This demonstrates the growing interest in leveraging LLMs not only for retrieval but also for semantic normalization and interoperability across food ontologies.

Furthermore, the development of specialized datasets, such as NutriBench \cite{hua2025nutribenchdatasetevaluatinglarge}, CafeteriaSA \cite{cenikj2022cafeteriasa} and CafeteriaFCD \cite{ispirova2022cafeteriafcd}, has facilitated the evaluation of LLMs in nutrition estimation tasks. NutriBench consists of meal descriptions annotated with macronutrient labels, providing a ground truth for assessing the performance of LLMs in estimating nutritional content from textual descriptions. CafeteriaSA represents the first annotated corpus of 500 scientific abstracts containing over 6,000 food entities linked to multiple semantic resources such as Hansard taxonomy, FoodOn, and SNOMED-CT. It serves as a gold standard for developing and evaluating food-domain named-entity recognition and linking methods from scientific text. Complementarily, CafeteriaFCD extends the FoodBase corpus by annotating food consumption data with semantic tags from four major resources Hansard, FoodOn, SNOMED-CT, and FoodEx2 enhancing semantic interoperability and supporting the creation of NLP models for food and dietary analysis applications.

\section{Methods}
This section provides a detailed description of the system and the evaluation framework and its constituent components used to assess the performance of LLMs. We outline the architecture of the RAG system, the dataset and database schema used for testing, and the procedure employed to evaluate query generation accuracy.

\subsection{System Overview}
\begin{figure}[tb]
	\centerline{\includegraphics[width=0.6\textwidth]{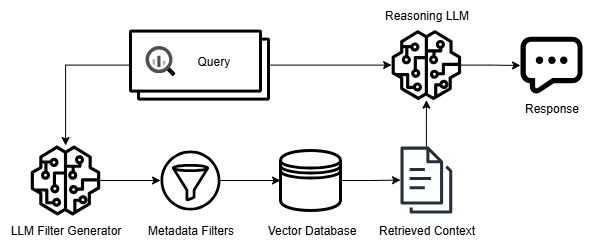}}
	\caption{System pipeline.}
	\label{fig:1}
\end{figure}

Our RAG system presented in Figure~\ref{fig:1} leverages Chroma \cite{chroma} for its open-source nature providing transparency and flexibility for specialized research environments, and its built-in filtering capabilities are directly relevant to our core mechanism for structured data retrieval.

The retrieval process begins by feeding each user query into an LLM that is specifically instructed to perform Chroma metadata filter generation where metadata includes component names, food names and food groups. Metadata filtering is a critical component that dramatically narrows the scope of the vector search, thereby vastly increasing the precision and accuracy of retrieval. This initial step is highly dependent on the LLMs ability to generate syntactically sound complex filters from natural language. Once the metadata filters are successfully generated, we query the vector database in a two-stage process: first using the metadata filters to restrict the search space, and then performing a traditional query similarity search within that restricted set to retrieve the final context.

In a complete RAG system, the retrieved context would be combined with the user’s original query and passed to a final LLM to generate the output response. In this study, we isolate and evaluate the retrieval component by comparing the retrieved context against a predefined \textit{ground truth} dataset to compute the retrieval accuracy score as illustrated in Figure~\ref{fig:3}.

\subsection{Data Collection and Preprocessing}

All of the data utilized in this study is sourced from the Slovenian Food Composition Database (FCDB), which is managed and maintained using the NutriBase system \cite{10.3389/fnut.2024.1503389}. NutriBase is a robust, integrated data management system designed to enhance the interoperability and reusability (FAIR principles) of food and nutrition-related data by linking food composition data with evidence-based knowledge. This system is crucial for ensuring the quality, harmonization, and trustworthiness of the underlying data and knowledge.

The FCDB contains two main types of food items:
\begin{itemize}
    \item \textbf{Branded foods:} These are commercially available products found in stores. For these foods, the database typically includes the name, food group and the big 8 components such as proteins, carbohydrates, total sugars, total fats (saturated and unsaturated), dietary fiber, and sodium. The data is often sourced from product labels and manufacturer information.
    
    \item \textbf{Generic foods:} These are standard generic foods that have been analyzed in the lab. Generic foods have very detailed composition data, including components such as micronutrients, vitamins, fatty acids, and trace elements (up to 366 components). This rich information allows for a highly granular representation of nutritional content.
\end{itemize}
\smallskip
Each food item, depending on its type, can therefore have a very different amount of data. In the preprocessing stage, we extracted all available attributes from each food item to construct the metadata for the Chroma vector database. For branded foods, this mostly included macronutrients and basic product information, whereas for generic foods, this included all available micro and macronutrient details. All component measurements were standardized to grams or kilocalories to ensure consistency across the dataset.

Following approaches from Pellegrini et al. \cite{pellegrini2021exploiting}, we converted the structured nutritional data into natural language descriptions. For instance:
\smallskip
\begin{quote}
\textit{Food item 'Cheese Provolon' belongs to the food group 'Cheeses', food group 'Cheeses'. Nutritional values per 100g: energy is 365.30 kcal, protein, total is 26.30 g, carbohydrates, total is 0.0 g, fat, total is 29.90 g, fibre, total dietary is 0.00 g, salt is 2.19 g.}
\end{quote}
\smallskip
In line with the approach of echo embeddings \cite{springer2025repetitionimproveslanguagemodel}, we repeated the \texttt{food\_group} name within each sentence to enhance its semantic representation in the embedding space. These text representations were then embedded using a large-scale embedding model (\texttt{gemini-embedding-001}) with an output dimensionality of 3072, and upserted into the Chroma vector database. This multimodal-inspired approach, combining structured data and natural language embeddings, allows flexible retrieval of similar foods, analogous to how FoodBERT and Food2Vec generate substitute recommendations \cite{pellegrini2021exploiting}. After preprocessing and embedding, our chroma database contained over 32,000 food items.

\subsection{Metadata Filter Generation}
The part of our system that contributes the most to our accuracy of retrieved context is the metadata filtering. The main goal is to accurately generate a Chroma metadata filter from natural language with correct syntax and component names. For that we use a normal non-finetuned model with an engineered prompt. We compared four different models: Google DeepMind (Gemini), OpenAI (GPT), Anthropic (Claude), and Mistral AI (Mistral). In the prompt we informed the model about all existing food component names, the syntactic rules about Chroma metadata filters and the exact format we want the response to be. If the LLM generates syntactically incorrect filters or makes a mistake at setting component names, we fallback to loose metadata filters which means that we instruct the LLM to only create filters for the food group component as it is the most important and distinguishable attribute of each food item.

\subsection{Retrieval Mechanism}
\label{sec:retriaval}
After the metadata filters are successfully obtained from the LLM, we execute the retrieval process.

First, we query the Chroma vector database using a two-step approach. 
Initially, the generated metadata filters are applied to drastically narrow the potential search space, 
i.e., the set of foods that satisfy explicit conditions such as \texttt{"protein > 12 g"}. 
Subsequently, a semantic similarity search is performed only within this filtered subset.  
Semantic search also ranks these compliant items based on their distance in the embedding space 
however, in our evaluation, the ranking itself is not considered. 

An example illustrating this process is shown below:
\begin{tcolorbox}[colback=white,colframe=gray!75!black,
  left=2mm, right=2mm, top=1mm, bottom=1mm, boxrule=0.5mm, arc=0.5mm]

\textbf{Query:}
\begin{lstlisting}[style=plaincode]
"Which foods have more than 12 g of protein?"
\end{lstlisting}

\textbf{Generated metadata filter:}
\begin{lstlisting}[style=plaincode]
{"protein, total": {"$gt": 12}}
\end{lstlisting}

\textbf{Returned item:}
\begin{lstlisting}[style=plaincode]
{
  "name": "Cheese Provolon",
  "food group": "Cheeses",
  "energy": 365.30,
  "protein, total": 26.30,
  "carbohydrates, total": 0.00,
  "fat, total": 28.90,
  "fibre, total dietary": 0.00,
  "salt": 2.19
}
\end{lstlisting}

\end{tcolorbox}

However, if the LLM generates syntactically incorrect filters, or if the filters contain erroneous component names, an exception is triggered, and the system initiates a graceful fallback sequence.
\begin{enumerate}
    \item \textbf{Loose Filtering Fallback:} We first attempt to generate loose metadata filters. These filters are engineered to focus only on the highly distinguishable attribute of the food item, the \texttt{food\_group} component (e.g., 'Cheeses'). This is based on the assumption that even a complex query often implicitly references a food category. If this succeeds, the semantic search is performed within this broader, food-group-restricted subset.
    \item \textbf{Pure Semantic Fallback:} If the loose filtering also fails, we abandon all metadata filtering and only retrieve context based on the semantic similarity of the query to all vectors in the database. This acts as a final fail-safe, ensuring some relevant context is retrieved, albeit with significantly lower precision than the filtered approaches.
\end{enumerate}

\subsection{Similarity Threshold for Context Selection}
\label{sec:threshold}
When metadata filtering is successful, the returned set is clearly defined by the filter conditions, and all matching items are returned regardless of their rank (up to the database limit).

However, when metadata filtering fails or is only partially successful, the system must rely solely on semantic similarity for retrieval. In such cases, ranking items by similarity alone is not sufficient, since the number of relevant results varies from query to query. Using a fixed top-k cutoff can either exclude relevant items or include many irrelevant ones, depending on the query’s specificity. To mitigate this issue, we introduced a similarity threshold that dynamically determines which items are considered relevant based on their similarity scores.

To find the optimal threshold, we calculated the cosine distance between all vectors in the database, where the distance $d$ is defined as:

\[
d(A, B) = 1 - \text{Cosine Similarity}(A, B) = 1 - \frac{A \cdot B}{\|A\| \, \|B\|}
\]

\begin{figure}[tb]
	\centerline{\includegraphics[width=0.6\textwidth]{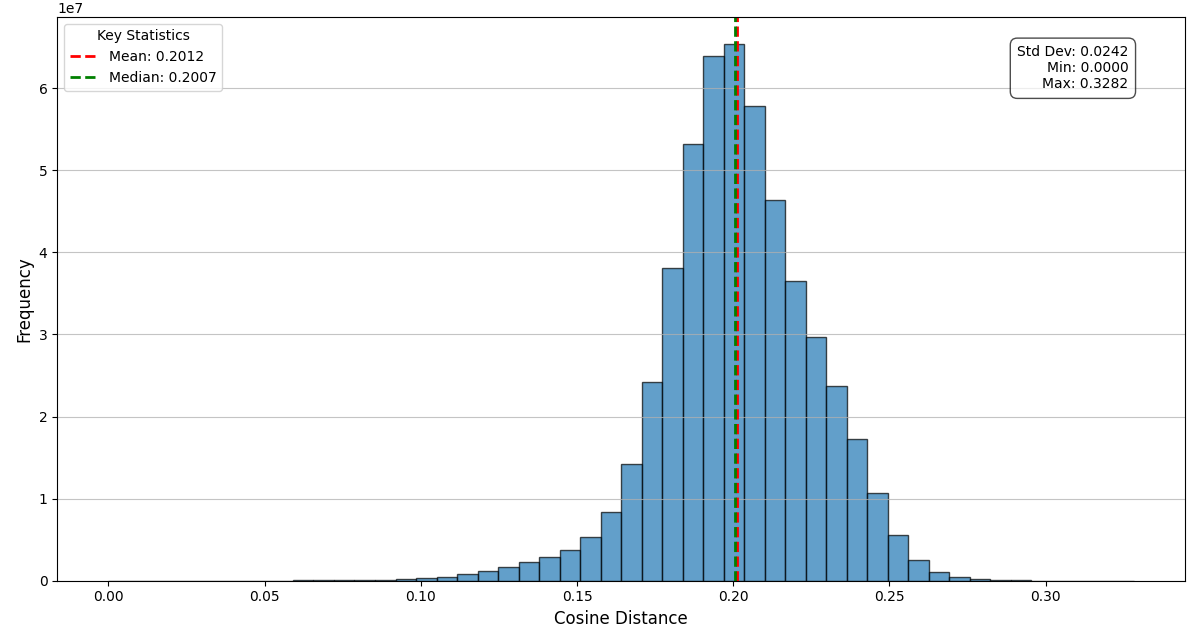}}
	\caption{Distances of all vectors between themselves in the database.}
	\label{fig:2}
\end{figure}

The distribution of these distances is shown in Figure \ref{fig:2}. We calculated the mean ($\mu$) and the standard deviation ($\sigma$) of this distribution, then tested the retrieval accuracy using three strategic similarity thresholds: $\mu - \sigma$, $\mu$, and $\mu + \sigma$. We selected the distance value threshold that maximized the retrieval score, ensuring that only vectors significantly similar to the query embedding are considered relevant in the absence of metadata constraints.

After the context is retrieved, it is sent to a reasoning model with the query.

\subsection{Evaluation of the System}
\begin{figure}[tb]
	\centerline{\includegraphics[width=0.6\textwidth]{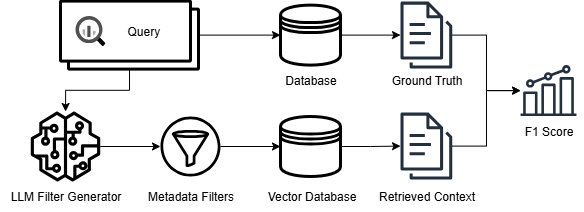}}
	\caption{Evaluation pipeline.}
	\label{fig:3}
\end{figure}

As previously mentioned, a rigorous evaluation of the system’s retrieval performance is necessary to assess how effectively LLMs can perform database querying. We constructed a test set consisting of 150 questions: 50 easy, 50 medium, and 50 hard, each with corresponding database queries established as the ground truth. We also ensured that every set of questions included some queries designed to yield no results, testing the system's ability to handle empty returns. The entire process was automated into an evaluation pipeline shown in the Figure~\ref{fig:3}, that included:

\begin{enumerate}
    \item \textbf{Filter Generation:} Creating structured metadata filters for each query using the LLMs under evaluation.
    \item \textbf{Vector Database Querying:} Executing the two-stage query (filter + similarity search).
    \item \textbf{Ground Truth Retrieval:} Manually identifying the correct set of food items that meet the specified query conditions. This human curated dataset serves as the benchmark for assessing the accuracy of the LLM generated queries.
    \item \textbf{Comparison and Scoring:} Comparing the retrieved set against the \textit{ground truth} using standard metrics.
\end{enumerate}

To evaluate the retrieval performance, we used the F1 score, which balances precision and recall.
Let \(G\) denote the set of ground truth items and \(R\) denote the set of retrieved items. Then:

\[
\text{True Positives (TP)} = |G \cap R|
\]
\[
\text{False Positives (FP)} = |R \setminus G|
\]
\[
\text{False Negatives (FN)} = |G \setminus R|
\]

The Precision, Recall, and F1 score are then defined as:

\[
\text{Precision} = \frac{\text{TP}}{\text{TP} + \text{FP}}
\]
\[
\text{Recall} = \frac{\text{TP}}{\text{TP} + \text{FN}}
\]
\[
\text{F1 Score} = 2 \cdot \frac{\text{Precision} \cdot \text{Recall}}{\text{Precision} + \text{Recall}}
\]

This metric ensures that the system is rewarded for retrieving relevant items while penalizing false positives and false negatives, providing a balanced measure of retrieval quality.

\subsection{Question Difficulty Categorization}

The 150 questions were categorized based on the complexity of the conditions required for metadata filter generation:

\textbf{Easy:} Consisted of 1 or 2 conditions.

\begin{itemize}
    \item \begin{quote}
        \textit{Which foods have more than 12 g of fats?}
    \end{quote}\vspace{4pt}
    \item \begin{quote}
        \textit{Which foods have less than 5 g of sugars and belong to the food group 'Fish'?}
    \end{quote}
\end{itemize}

\textbf{Medium example queries:} Consisted of 3 to 4 conditions, including nested AND/OR logic and range queries.

\begin{itemize}
    \item \begin{quote}
        \textit{Which foods have more than 0.5 g of potassium, more than 0.2 g of magnesium, more than 0.01 g of vitamin C, and less than 5 g of fats?}
    \end{quote}\vspace{4pt}
    \item \begin{quote}
        \textit{Which foods from the food group 'Fresh beef' or 'Dry fruits' have more than 10 g of proteins, less than 2 g of sugars, less than 15 g of fats, and less than 5 g of carbohydrates?}
    \end{quote}\vspace{4pt}
    \item \begin{quote}
        \textit{Which foods have proteins between 30 and 35 g?}
    \end{quote}
\end{itemize}

\textbf{Hard example queries:} Contained conditions requiring advanced reasoning, such as comparative questions or aggregate calculations.
\begin{itemize}
    \item \begin{quote}
        \textit{Which foods from the food group 'Chicken meat' have more proteins than cholesterol?}
        \end{quote}
    \vspace{4pt}
    \item \begin{quote}
        \textit{Which foods have a sum of proteins and fats greater than 80 g?}
        \end{quote}
\end{itemize}

\subsection{Retrieval Mechanism Testing}

Hard queries were constructed to evaluate how well the system can leverage loose filtering and pure semantic fallback when strict metadata filter generation is difficult. This tests the efficacy of the system's fallback mechanisms, specifically the loose filtering and pure semantic retrieval.

We evaluated a total of 150 queries across four LLMs, \textit{Gemini-2.0-Flash}, \textit{GPT-4o}, \textit{Mistral Medium 3}, and \textit{Claude-Sonnet-4} and tested each model under three different similarity thresholds (five full runs per model per threshold) to ensure stability and robustness of the results. These models were selected as they are widely recognized as some of the most capable general-purpose LLMs currently available.

\section{Results}
Table~\ref{tab:evaluation} reports the average F1 scores (computed over five independent runs) for all evaluated models across the three difficulty categories and for each of the three similarity thresholds (\(\mu - \sigma\), \(\mu\), and \(\mu + \sigma\)). 

The single highest average F1 observed in our experiments was \(0.450\), obtained by \textit{Claude} at the middle similarity threshold (\(\mu \approx 0.613\)). However, when averaging performance across all four models, the more restrictive threshold \(\mu - \sigma \approx 0.539\) produced the highest mean F1 for the Hard category (mean F1 \(\approx 0.424\)), indicating that a more restrictive acceptance criterion can improve overall robustness for semantic fallback retrieval.

In the \textbf{Easy} category, all models achieved very high performance (\(F1 > 0.999\)) across all thresholds, confirming that when metadata filters are generated correctly the retrieval process reliably returns the relevant food items.

A minor technical artifact was observed: in cases where a metadata filter matched several thousand food items, Chroma occasionally failed to retrieve a small subset of expected results despite them satisfying the filter. This appears to be a limitation in the vector database’s internal index handling rather than an issue with filter generation or data representation. As the issue affected all models equally and only in large-result scenarios, it does not affect the comparative conclusions.

In the \textbf{Medium} category, all models maintained high performance across the three thresholds. At \(\mu\) both \textit{Gemini} and \textit{Claude} reached \(F1=1.000\), followed by \textit{Mistral} (\(0.998\)) and \textit{GPT} (\(0.990\)). This indicates that moderately complex filter structures involving compound logical operators and range constraints remain within the capabilities of current LLMs.

In the \textbf{Hard} category, where queries require comparative or aggregate reasoning that challenge strict metadata filter generation, performance decreased for all models. At the \(\mu - \sigma\) threshold (\(\approx 0.539\)) the average F1 scores were: \textit{Gemini} \(0.445\), \textit{Claude} \(0.431\), \textit{GPT} \(0.396\), and \textit{Mistral} \(0.425\). Recall that the largest average score across all thresholds remained \textit{Claude}'s \(0.445\) at \(\mu \approx 0.613\).

\begin{table}[tbh]
    \footnotesize
    \caption{Average F1 scores.}
    \begin{center}
    \resizebox{\columnwidth}{!}{%
    \begin{tabular}{ccccc}
        \toprule
        \textbf{Model} & \textbf{Threshold} & \textbf{Easy} & \textbf{Medium} & \textbf{Hard} \\
        \midrule
        Gemini  & \multirow{4}{*}{0.539} & 0.999($\pm$0.000) & 1.000($\pm$0.000) & 0.445($\pm$0.009) \\
        Claude  &                        & 0.999($\pm$0.000) & 1.000($\pm$0.000) & 0.431($\pm$0.025) \\
        GPT     &                        & 0.999($\pm$0.000) & 0.994($\pm$0.008) & 0.396($\pm$0.012) \\
        Mistral &                        & 0.999($\pm$0.000) & 0.998($\pm$0.001) & 0.425($\pm$0.013) \\
        \midrule
        Gemini  & \multirow{4}{*}{0.613} & 0.999($\pm$0.000) & 1.000($\pm$0.000) & 0.420($\pm$0.015) \\
        Claude  &                        & 0.999($\pm$0.000) & 1.000($\pm$0.000) & \textbf{\color{Green}0.450}($\pm$0.013) \\
        GPT     &                        & 0.999($\pm$0.000) & 0.990($\pm$0.008) & 0.389($\pm$0.017) \\
        Mistral &                        & 0.999($\pm$0.000) & 0.998($\pm$0.000) & 0.427($\pm$0.006) \\
        \midrule
        Gemini  & \multirow{4}{*}{0.686} & 0.999($\pm$0.000) & 1.000($\pm$0.000)  & 0.443($\pm$0.023) \\
        Claude  &                        & 0.999($\pm$0.000) & 1.000($\pm$0.000)  & 0.419($\pm$0.028) \\
        GPT     &                        & 0.999($\pm$0.000) & 0.994($\pm$0.008)  & 0.373($\pm$0.012) \\
        Mistral &                        & 0.999($\pm$0.000) & 0.998($\pm$0.000)  & 0.418($\pm$0.011) \\
        \bottomrule
    \end{tabular}%
    }
    \label{tab:evaluation}
    \end{center}
\end{table}

\section{Discussion}
The evaluation demonstrated consistently high retrieval performance for both the \textbf{Easy} and \textbf{Medium} question categories. All models achieved very high average F1 scores (\(>0.999\)) on Easy queries, and both \textit{Gemini} and \textit{Claude} reached \(F1=1.000\) on Medium queries, closely followed by \textit{Mistral} (\(0.998\)) and \textit{GPT} (\(0.994\)). This showcases that, even without fine-tuning, an open-source LLM such as \textit{Mistral} can serve as a highly reliable metadata filter generator within a RAG pipeline.

Performance on the \textbf{Hard} category across the three tested similarity thresholds showed two complementary findings. First, the single best average retrieval run was produced by \textit{Claude} at the middle threshold (\(\mu \approx 0.613\)) with \(F1=0.450\). Second, averaging across models, the most restrictive threshold \(\mu-\sigma \approx 0.539\) yielded the highest mean F1 (\(\approx 0.424\)), indicating that a more restrictive acceptance criterion for semantic similarity tends to improve overall robustness of fallback retrieval. The Hard queries were designed to exceed the filter-generation capabilities of the Chroma vector database by requiring comparative reasoning or aggregate calculations. The fact that on average \textit{Claude} recovered over 44\% of relevant items in the best run shows that partial retrieval is achievable even when exact filter construction fails. Further research is needed to systematically improve retrieval performance under such constraints.

These results validate that LLM-driven metadata filtering can be an effective mechanism for accessing structured nutritional data. This approach enables domain experts, such as nutritionists and dietitians, to access database information through natural language queries without the need for technical mediation. However, its reliability is largely confined to simple and well-structured questions. As query complexity increases, the retrieval accuracy of LLMs declines sharply, representing a significant limitation in realistic, real-world use cases.

\subsection{Limitations}
Despite the system's high overall accuracy, we identify a few key limitations: 

We used and tested only one vector database, Chroma. The high retrieval scores are slightly tempered by a minor observed anomaly where the system occasionally failed to return the complete set of vectors when the metadata filter matched several thousands of entries. In order to fully address this implementation challenge future work should involve a comparative study with other vector databases to assess index efficiency for large, filtered result sets.

Language and Generalizability: The data collection and all evaluation queries were conducted entirely in the Slovene language. While this may limit the immediate generalization of the results, it also serves as a positive indicator of the tested LLMs' capabilities. The high performance demonstrates that non fine-tuned models possess sufficient cross-lingual and multilingual understanding to accurately generate structured queries and perform semantic search in a resource-scarce language like Slovene.

Another important direction for future research is the comparison of model generations. Preliminary testing revealed that \textit{Gemini-2.5-Pro} performed noticeably worse than its predecessor \textit{Gemini-2.0-Flash}, despite being a newer release. This finding underscores the need for systematic evaluation of newer model iterations to better understand the trade-offs introduced by architectural or alignment changes.
Additionally, future studies should examine the price–performance ratio of different LLMs, as the cost of inference plays a crucial role in determining the practical scalability and deployment feasibility of such systems.

\ack{The authors acknowledge the support of the Slovenian Research Agency through program grant No. P2-0098 and project grant No. GC-0001. We also acknowledge the support of the EC/EuroHPC JU and the Slovenian Ministry of HESI via the project SLAIF (grant number 101254461).}

\appendix

\end{document}